\documentclass{article}
\usepackage{xcolor}
\usepackage{multirow} 
 \usepackage{booktabs}
\usepackage{todonotes}
\usepackage{amssymb}
\usepackage{spconf,amsmath,graphicx,hyperref}

\title{Integrating Language Models into Listened and Imagined Speech Decoding from MEG}
\name{Maryam Maghsoudi$^{\star}$ \qquad Sai Samrat Kankanala$^{\dagger}$  \qquad Shihab A. Shamma$^{\star}$ \qquad Sriram Ganapathy$^{\dagger}$}
\address{$^{\star}$Department of Electrical and Computer Engineering, University of Maryland, College Park, MD, USA \\
	$^{\dagger}$LEAP Lab, Electrical Engineering, Indian Institute of Science, Bengaluru, India.}
\begin{document}
%
\maketitle
\begin{abstract}
Decoding imagined speech is an important goal for brain-computer interfaces but remains challenging due to weak neural responses, low signal-to-noise ratio, and limited imagined-speech datasets. Language models provide strong contextual cues for text prediction, but how much they can help neural decoding and whether their contribution differs for decoding perceived and imagined speech remains unclear. To investigate this, we use a paired listened-imagined MEG dataset and incorporate language-model information at two stages. First, we train a contrastive neural decoder that aligns MEG representations with acoustic and contextual language representations, improving cross-subject word decoding for both listened and imagined speech. Second, at inference, we introduce a neural-constrained beam-search framework that combines neural evidence with language-model next-word probabilities. We find that imagined-speech decoding benefits more from the language model than listened-speech decoding. For Imagined speech, the best-performing balance between neural and language-model evidence shifts toward the language model, and the gain over neural-only decoding is larger. Together, these results suggest that language priors are most useful when neural evidence is weaker, making them particularly valuable for imagined-speech BCIs.
\end{abstract}


\section{Introduction}
\label{sec:intro}
Decoding imagined speech from brain activity could enable communication for people who are unable to speak or vocalize \cite{moses2021neuroprosthesis,willett2023high}. However, imagined speech remains considerably harder to decode than perceived speech. Neural responses during imagination are weaker and less reliably time-locked than responses evoked by an external speech signal \cite{martin2014decoding}. Imagined-speech datasets are also relatively small, in part because imagination has no directly observable output and its timing is difficult to measure precisely \cite{alharbi2024decoding,maghsoudi2025convolutional}. In contrast, decoding perceived speech has advanced rapidly, with recent work reconstructing speech and language from both invasive and non-invasive neural recordings \cite{defossez2023decoding,akbari2019towards}.

Prior work has used language models to decode speech from neural data by incorporating linguistic context and prior knowledge \cite{tang2023semantic}. Language priors have been shown to support the reconstruction of language from both invasive and non-invasive neural recordings \cite{willett2023high, card2024accurate, levy2025brain}. Their application to imagined speech, however, remains limited. Imagined speech has weaker neural evidence and is inherently noisier than listened speech. It remains unclear how the benefit of a language prior scales with the reliability of the underlying neural evidence, and in particular, whether listened and imagined speech benefit from it differently.

\begin{figure*}[t]
  \centering
  \includegraphics[width=\linewidth]{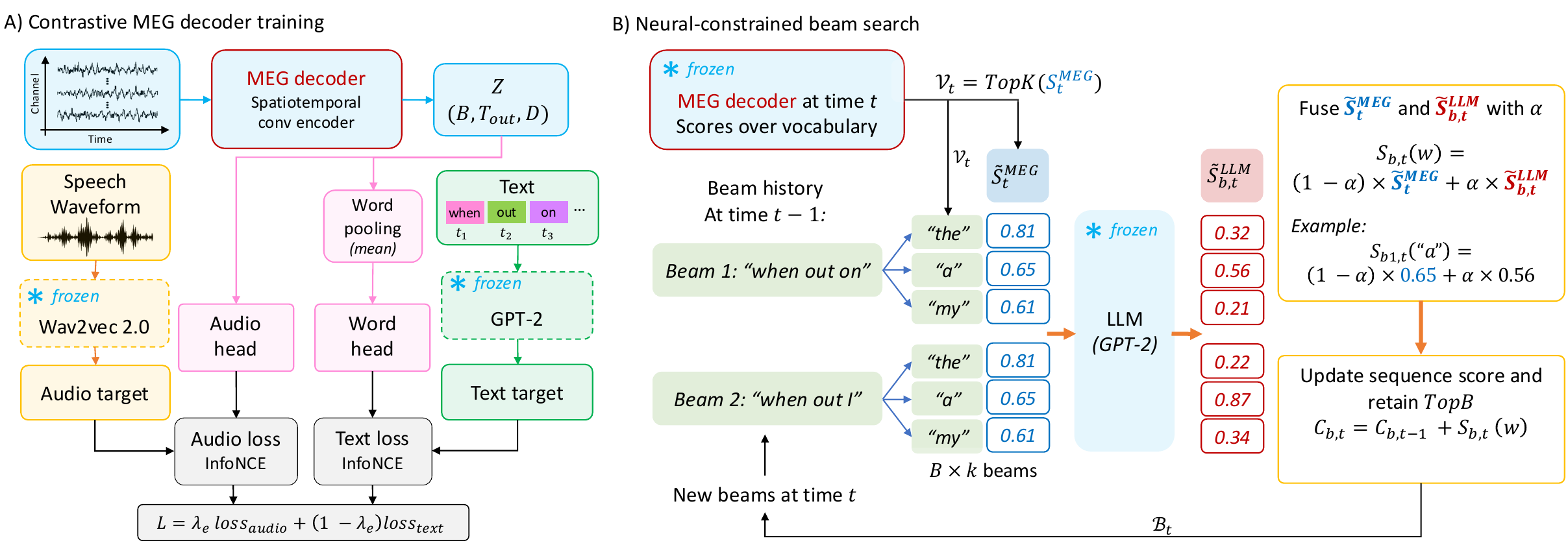}
  \caption{(A) Training of the contextual contrastive decoder. A shared MEG encoder is trained with two InfoNCE objectives: frame-level alignment to wav2vec~2.0 and word-level alignment to GPT-2 hidden states, combined with weight $\lambda_e$. (B) Neural-constrained beam search at inference. At each position $t$, the frozen MEG decoder selects the top-$k$ candidate words $\mathcal{V}_t$, and the frozen LLM scores each candidate given each beam's decoded history. The normalized MEG and LLM scores are fused with weight $\alpha$ and added to the beam's cumulative score, and the $B$ highest-scoring sequences are retained.}
  \label{fig:decoder}
\end{figure*}

In this work, we ask: \textit{how does the contribution of a language model change when decoding listened versus imagined speech?} To answer this question, we leverage a paired listened--imagined MEG dataset introduced in \cite{maghsoudi2026zero}, in which the same participants either listen to or imagine the same poetic stimuli. 
 Building on the imagined-to-listened mapping framework of \cite{maghsoudi2026zero},
we incorporate language-model information in two ways: during neural decoder training, where contrastive learning aligns MEG with acoustic and contextual language representations, and during inference, where neural predictions are combined with language-model next-word probabilities. By varying the relative contribution of neural and language-model evidence, we examine how their optimal balance changes between listened and imagined speech.

Our main contributions are: (i) We introduce a contrastive neural decoder jointly trained against acoustic and contextual language representations, and show that it substantially improves decoding accuracy for both listened and imagined speech. (ii) We develop a neural-constrained beam-search framework that explicitly controls the relative contribution of neural and language-model evidence during decoding, enabling a direct comparison of their contributions across listened and imagined speech. (iii) We show that the optimal balance between neural and language-model evidence differs between the two conditions: imagined speech achieves its best decoding performance with a greater language-model contribution and exhibits a larger improvement over neural-only decoding. 

These results suggest that language priors become increasingly valuable as neural evidence becomes less reliable, highlighting their potential for imagined-speech decoding.

\section{Methods}
\subsection{Dataset}
\label{sec:data}
We use the paired listened–imagined MEG dataset of~\cite{maghsoudi2026zero}, comprising 17 participants either listened to or imagined two poems, each 27 seconds long and repeated 10 times per condition. Four subjects were excluded due to data-quality issues, leaving 13 for analysis. We adopt the imagined-to-listened neural mapping introduced in~\cite{maghsoudi2026zero}. Under leave-one-subject-out (LOSO), the mapping $M_{-s}$ is trained without any data from the held-out subject $s$, and transforms an imagined MEG trial $X_s^{I}$ into an estimate $\hat{X}_s^{L}$ of the corresponding listened response. We use all six mapping variants from~\cite{maghsoudi2026zero} and average imagined-speech results across them.
We train our own decoder on listened trials only (Section~\ref{sec:decoder}) for each held-out subject, following the same LOSO protocol, so that no stage of the pipeline sees the held-out subject.

\subsection{Contextual Contrastive Decoder}
\label{sec:decoder}
To align MEG with both the acoustic and linguistic information in the stimulus, we train a neural encoder on listened trials with two contrastive objectives (Fig.~\ref{fig:decoder}A). Given a continuous MEG trial $X$, the encoder $E_\theta$, a $1{\times}1$ spatial convolution followed by a causal dilated temporal convolutional network (TCN), produces latent representations $Z = E_\theta(X) \in \mathbb{R}^{T_{\text{out}} \times D}$ at approximately 50~Hz.

\textbf{Acoustic objective.} To align MEG with speech at the frame level, we project $Z$ with an audio head and apply an InfoNCE loss~\cite{oord2018representation} against hidden states of a frozen wav2vec~2.0 model~\cite{baevski2020wav2vec} (layer $\ell_a$), at corresponding time points.

\textbf{Language objective.} To align MEG with linguistic context at the word level, we mean-pool $Z$ over each word's onset--offset window, project it with a word head, and apply an InfoNCE loss against hidden states of a frozen GPT-2 ~\cite{radford2019language} (layer $\ell_w$). Targets are computed from the poem text, so each word representation incorporates its preceding context. Positives are batch samples sharing the same poem and word position across subjects and sessions, used only to identify matching GPT-2 targets for the loss, never provided as input to the encoder, which receives only the raw MEG trial $X$.

\textbf{Joint training.} The two objectives are optimized with 
$
\mathcal{L}_{\mathrm{train}} = \lambda_e \mathcal{L}_{\mathrm{audio}} + (1-\lambda_e)\mathcal{L}_{\mathrm{LLM}},$
where $\lambda_e$ follows a cosine anneal~\cite{loshchilov2016sgdr,bengio2009curriculum} from $0.8$ to $0.2$ over the first 15 epochs ($e$ is the epoch). This schedule initially emphasizes acoustic alignment, which provides fine-grained temporal supervision, and gradually shifts the training emphasis toward contextual language representations. We select $\ell_a = 6$ and $\ell_w = 8$ by grid search. These intermediate layers are also consistent with prior evidence that middle layers of self-supervised speech and language models best align with brain responses to speech~\cite{vaidya2022self,caucheteux2021gpt}.
\begin{table}[t]
\centering
\caption{Word decoding performance for listened and mapped imagined MEG (mean $\pm$ SEM across held-out subjects). Imagined results are averaged across six imagined-to-listened mappings. Shuffled: decoder trained with shuffled neural--target pairs.}
\label{tab:decoder_results}
\small
\setlength{\tabcolsep}{3pt}
\begin{tabular}{@{}llcc@{}}
\toprule
\textbf{Input} & \textbf{Decoder (LOSO)} & \textbf{Top-1 (\%)} & \textbf{Top-5 (\%)} \\
\midrule
\multirow{4}{*}{Listened}
& Previous~\cite{maghsoudi2026zero} & $1.44 \pm 0.16$ & $7.45 \pm 0.60$ \\
& Proposed & $\mathbf{14.17 \pm 0.81}$ & $\mathbf{46.96 \pm 1.56}$ \\
& Proposed, shuffled & $4.12 \pm 0.10$ & $18.41 \pm 0.18$ \\
\midrule
\multirow{4}{*}{Mapped imag.}
& Previous~\cite{maghsoudi2026zero} & $1.26 \pm 0.09$ & $6.71 \pm 0.21$ \\
& Proposed & $\mathbf{7.05 \pm 0.35}$ & $\mathbf{31.14 \pm 0.97}$ \\
& Proposed, shuffled & $3.93 \pm 0.08$ & $18.69 \pm 0.14$ \\
\bottomrule
\end{tabular}
\end{table}

\subsection{Language-Model Fusion and Neural-Constrained Beam Search}
\label{sec:beam}
The trained decoder scores each word from neural evidence alone, without knowing which words are plausible given the preceding context. To add this knowledge, we combine its predictions with the next-word probabilities of a language model at inference (Fig.~\ref{fig:decoder}B), with no parameter updates, for both listened and mapped imagined responses. At each word position $t$, the encoder and word head produce an embedding $z_t$, and each word $w$ in the decoding vocabulary $\mathcal{V}$ receives a neural score $S^{\mathrm{MEG}}_t(w) = \max_{j: w_j = w} \cos(z_t, h_j)$, where $h_j$ is the GPT-2 representation at position $j$ of the two poems. Because the neural and language-model scores have different scales, we independently z-score each set of scores across the vocabulary at every word position, yielding $\tilde{S}^{\mathrm{MEG}}_t$ and $\tilde{S}^{\mathrm{LLM}}_t$. The two normalized scores are then combined by weighted interpolation,
\begin{equation}
S_t(w \mid \text{context}) = (1-\alpha)\,\tilde{S}^{\mathrm{MEG}}_t(w) + \alpha\,\tilde{S}^{\mathrm{LLM}}_t(w \mid \text{context}),
\label{eq:fusion}
\end{equation}
where $\alpha \in [0,1]$ controls the relative contribution of neural and linguistic information.

\textbf{Teacher-forced fusion.} To quantify the maximum benefit available from ground-truth linguistic context within our fusion framework,
GPT-2 conditions on the true preceding words and scores all of $\mathcal{V}$, and we select $\arg\max_w S_t(w \mid w_{1:t-1})$ independently at each position.

\textbf{Neural-constrained beam search.} At inference, the true history is unavailable, and language-model errors can compound across steps. We therefore let MEG gate the candidate set: at each position, GPT-2 scores only the top-$k$ words from MEG, $\mathcal{V}_t = \mathrm{TopK}_{w \in \mathcal{V}}\, S^{\mathrm{MEG}}_t(w)$, conditioned on each beam's own decoded history $\hat{w}^{(b)}_{1:t-1}$. Each fused score from Eq.~\eqref{eq:fusion} is added to the beam's cumulative score, and the $B$ highest-scoring sequences are retained.

\subsection{Evaluation, Fusion Sweep, and Controls}
\textbf{Evaluation.}
All models are evaluated with LOSO cross-validation (Section~\ref{sec:data}). We report word accuracy as mean $\pm$ SEM across held-out subjects: top-1 and top-5 for the contrastive decoder, and top-1 for fused sequences. To quantify what the LM adds, we report the relative gain over MEG-only decoding and compare conditions with paired tests across subjects.

\textbf{Fusion sweep.}
We sweep $\alpha$ from 0 to 1 across beam widths $B$ and candidate-set sizes $k$. At $\alpha=0$ candidates are ranked by MEG scores alone, so the output reduces to the decoder's top-1 prediction; at $\alpha=1$, MEG still selects $\mathcal{V}_t$, but ranking within it comes entirely from the LLM. For quantitative comparisons, $\alpha$ is selected within each LOSO fold and applied to the held-out subject.

\textbf{Controls.}
For the decoder, \emph{Zero MEG} replaces $X$ with an all-zero trial, and \emph{Shuffled MEG} trains and evaluates with shuffled neural--target correspondence. For fusion, the \emph{random-candidate null} replaces the MEG top-$k$ set with $k$ random vocabulary words, retaining their MEG scores, and \emph{LLM only} runs beam search with GPT-2 alone over the full vocabulary, identical for both conditions.

\begin{figure}[t]
  \centering
  \includegraphics[width=\linewidth]{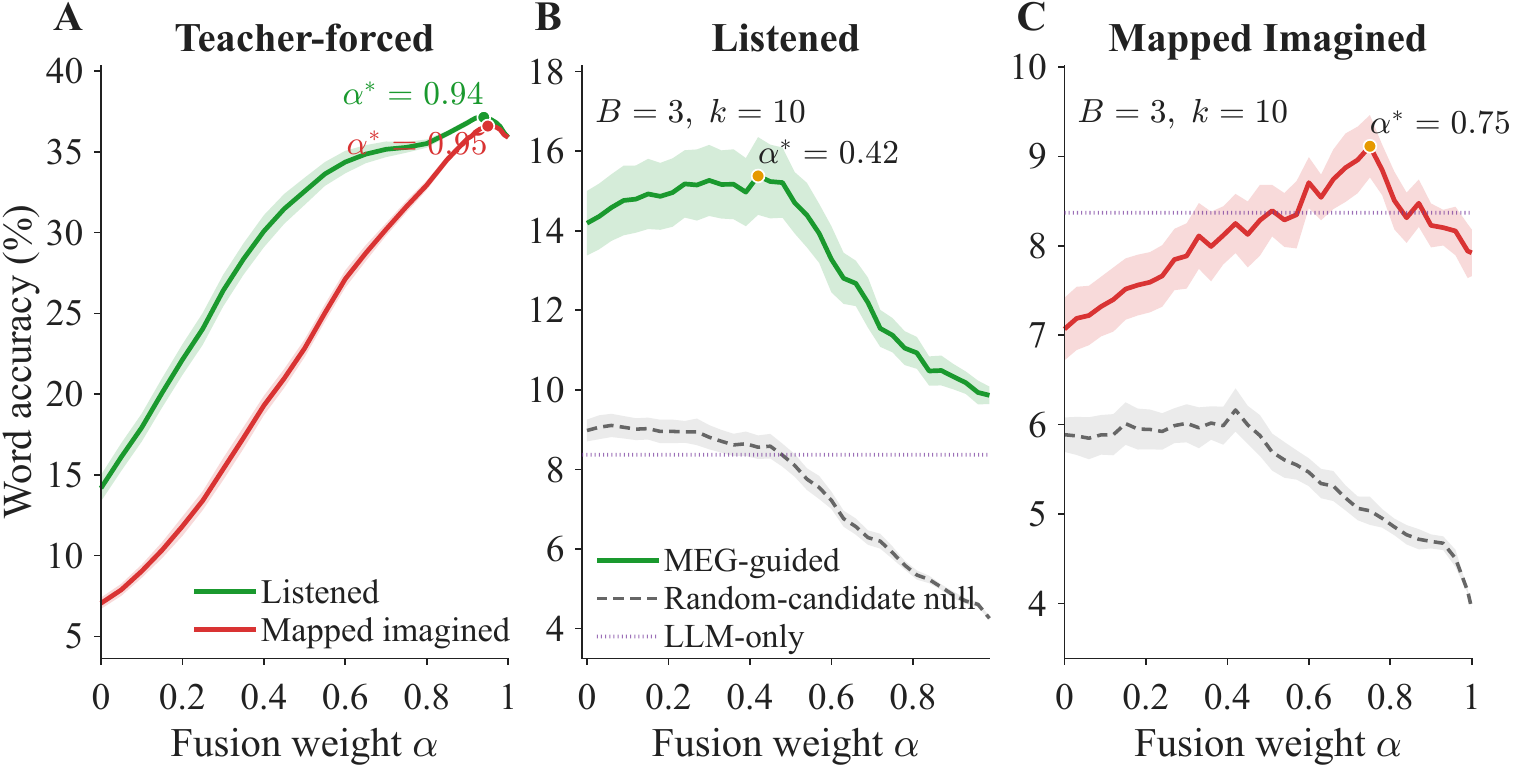}
  \caption{Word accuracy versus fusion weight $\alpha$. (A)~Teacher forcing, for listened and mapped imagined MEG. (B,C)~Neural-constrained beam search at B=3 and k=10 for listened (B) and mapped imagined (C). Solid curves show MEG-guided candidates, dashed curves the random-candidate null, and the dotted line LLM-only beam search over the full vocabulary. Shaded regions denote SEM across subjects, and markers indicate the $\alpha$ that maximizes the mean curve. Y-axes differ across panels. Under beam search, both conditions peak at an intermediate $\alpha$.}
  \label{fig:beam_fusion}
\end{figure}

\section{Results}
\begin{figure}[t]
  \centering
  \includegraphics[width=\linewidth]{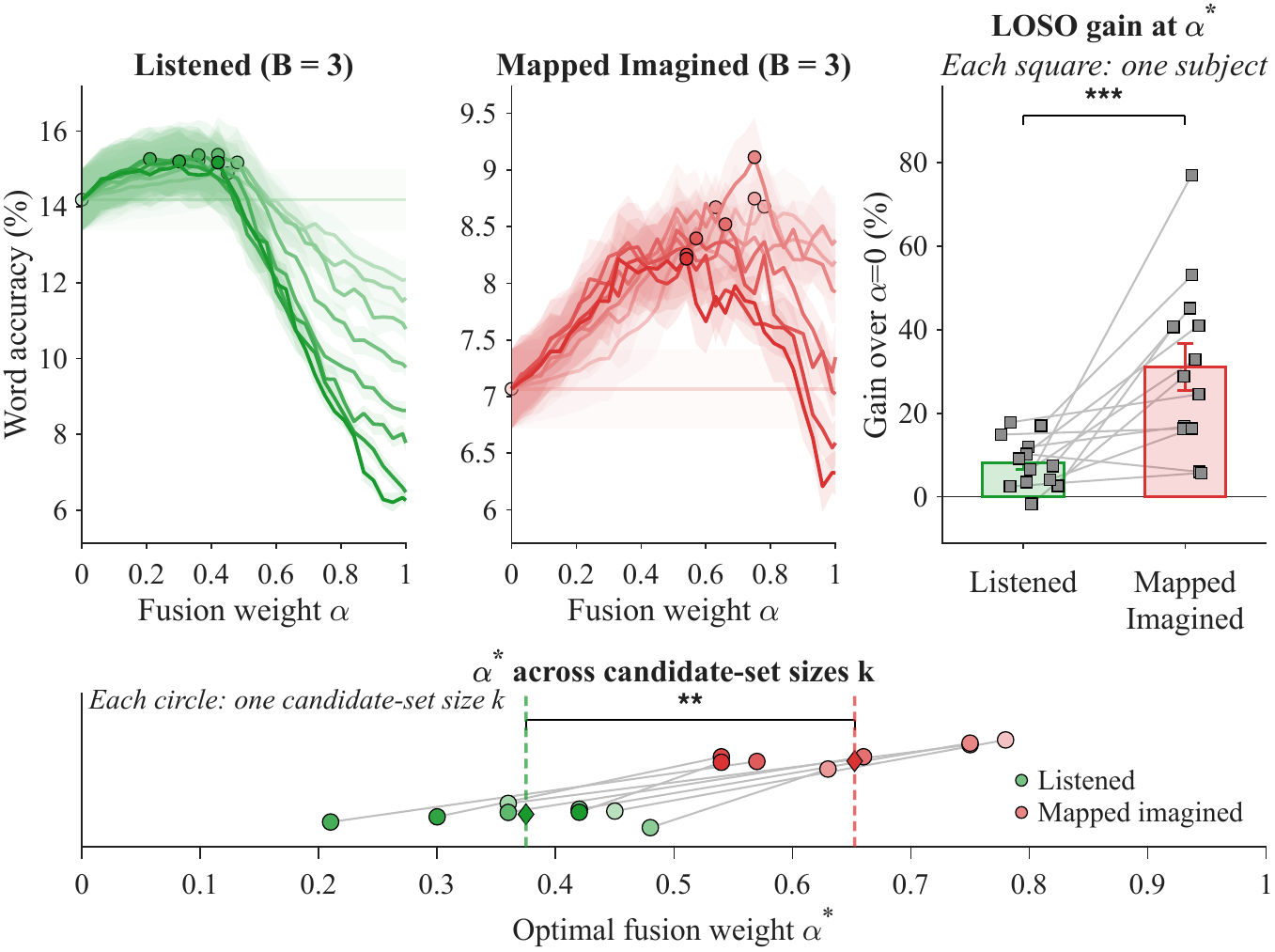}
  \caption{Comparison of the neural--language balance for listened and mapped imagined MEG. Top left and middle: word accuracy across $\alpha$ for different candidate-set sizes $k$ at $B=3$, with lighter-colored curves for smaller $k$; shaded regions denote SEM across subjects, markers indicate $\alpha^*$ for each $k$, and the y-axes differ across panels. Top right: relative gain over MEG-only ($\alpha=0$) at the selected $\alpha^*$, with one square per held-out subject, lines linking the same subject across conditions, and bars showing mean $\pm$ SEM. Bottom: $\alpha^*$ for each $k$, with one circle per $k$ and dashed lines marking the mean per condition. Mapped imagined responses show a higher optimal $\alpha$ and a larger gain than listened responses.}
  \label{fig:alpha_k}
\end{figure}

\textbf{Contextual Decoder Improves Cross-Subject Word Decoding.} The proposed decoder substantially improves cross-subject decoding over the previous decoder for listened MEG, and this improvement transfers to mapped imagined responses, exceeding its controls for all six mappings at both top-1 and top-5 (all $p<0.001$; Table~\ref{tab:decoder_results}). On all-zero input, its accuracy falls to chance level. The acoustic objective appears to prevent a shortcut: with the language objective alone, the decoder scores above chance even on zero input by defaulting to frequent-word targets, whereas frame-level acoustic alignment encourages it to track the neural input. Also, accuracy for mapped imagined responses is about half that for listened responses at top-1, so the two conditions provide neural evidence of different strength.

\textbf{Inference-Time Fusion Balances Neural and Language Evidence.} To measure how much linguistic context can contribute in principle, we first condition the language model on the ground-truth preceding words (teacher forcing, Fig.~\ref{fig:beam_fusion}A). In both conditions, accuracy rises with $\alpha$ until $\alpha$ is close to 1, and the two conditions converge at high $\alpha$. With the correct history, the language model dominates ranking and the neural evidence adds little. This analysis therefore bounds the benefit of context but does not let MEG contribute, and it assumes a history that is unavailable at inference.

We therefore evaluate neural-constrained beam search, where the language model conditions on its own decoded hypotheses and ranks only candidates proposed by MEG (Fig.~\ref{fig:beam_fusion}B,C; $B=3$, $k=10$). Unlike teacher forcing, both conditions show an intermediate optimum in $\alpha$, and joint scoring improves over MEG-only ranking in both ($p<0.001$). Both conditions also stay above the random-candidate null across $\alpha$, so neural candidate selection matters in each. For listened MEG, accuracy stays above the LLM-only baseline at every $\alpha$. For mapped imagined MEG, it exceeds the LLM-only baseline in a window near the optimum.

The endpoints of the sweep show a further difference between the conditions (Fig.~\ref{fig:beam_fusion}B,C). For listened MEG, ranking candidates by MEG scores alone ($\alpha=0$) outperforms ranking them by the language model alone ($\alpha=1$). For mapped imagined MEG, this order is reversed. In neither condition is a single source optimal, since accuracy peaks between the endpoints, at a different $\alpha$ in each. We next examine how this balance shifts between conditions.

\textbf{Imagined Speech Shifts the Optimal Balance Toward the Language Model.} To compare how the two conditions balance neural and language evidence, we sweep $\alpha$ across candidate-set sizes $k$ and beam widths $B$. Results were consistent across $B$, so we show $B=3$ (Fig.~\ref{fig:alpha_k}, top left and middle). In both conditions, accuracy peaks at an intermediate $\alpha$ for every $k>1$, and the drop toward $\alpha=1$ becomes steeper as $k$ grows. With small $k$, MEG constrains decoding by limiting the language model to a few neurally selected words, so neural information remains present even at $\alpha=1$. As $k$ grows, this constraint weakens and the language model can choose among more candidates, so removing the relative MEG scores becomes more costly. MEG therefore contributes in two ways: by selecting candidates and by ranking the words within that set.

The optimal fusion weight $\alpha^*$ is higher for mapped imagined than for listened MEG across $k$ (Fig.~\ref{fig:alpha_k}, bottom; $p<0.01$). Thus, mapped imagined responses favor a larger language-model contribution before performance begins to decline.

To quantify the contribution of language model in each condition, we compare the relative gain over MEG-only ranking ($\alpha=0$) at the selected $\alpha$ (Fig.~\ref{fig:alpha_k}, top right). The gain is larger for mapped imagined than for listened MEG (paired test across subjects, $p<0.001$). Since mapped imagined MEG provides weaker neural evidence (Table~\ref{tab:decoder_results}), these results are consistent with language-model information becoming more valuable as neural evidence becomes less reliable.

\section{Discussion and Limitations}
Our results show that the optimal balance between neural and linguistic evidence differs across conditions. Both listened and imagined speech peak at an intermediate language-model weight, but imagined speech favors a larger language contribution and gains more over neural-only decoding. This is consistent with language priors becoming more useful as neural evidence weakens. The teacher-forcing results further support this interpretation: with the true history, increasing language weight helps monotonically, whereas autoregressive decoding introduces language-model errors and a trade-off between neural and linguistic information.

Our evaluation is limited to two poems, a closed vocabulary, 13 trained musicians, and known word boundaries. LOSO tests generalization across subjects, but not to unseen stimuli or continuous speech. The reported results use GPT-2; we also observe the same trend with SmolLM \cite{allal2025smollm2}, suggesting that the effect is not specific to one language model.

These findings suggest that future speech BCIs may benefit from adapting language context to the reliability of neural predictions, relying more on neural evidence when strong and more on linguistic context when uncertain.

\vfill\pagebreak

\bibliographystyle{IEEEbib}
\bibliography{strings,refs}

\end{document}